\PassOptionsToPackage{numbers}{natbib}
\documentclass[conference]{IEEEtran}
\IEEEoverridecommandlockouts
\usepackage{cite}
\usepackage{amsmath,amssymb,amsfonts}
\usepackage{algorithmic}
\usepackage{graphicx}
\usepackage{textcomp}
\usepackage{stfloats}
\usepackage{float}
\usepackage{xcolor}
\usepackage{booktabs}
\usepackage{multirow}
\usepackage{url}
\usepackage{natbib}
\usepackage{amsmath,amssymb,bm}
\usepackage{tikz}
\usetikzlibrary{calc,positioning,arrows.meta,backgrounds,fit}
\usepackage{booktabs}
\usepackage{xcolor}

\usepackage{tikz}
\usetikzlibrary{shapes,arrows,positioning,calc}
\usepackage{amsmath}
\usepackage{amssymb}

\def\BibTeX{{\rm B\kern-.05em{\sc i\kern-.025em b}\kern-.08em
    T\kern-.1667em\lower.7ex\hbox{E}\kern-.125emX}}
\begin{document}

\title{PSTNet: Physically-Structured Turbulence Network
}

\author{
  \IEEEauthorblockN{Boris Kriuk}
  \IEEEauthorblockA{\textit{Department of Computer Science \& Engineering} \\
    \textit{Hong Kong University of Science and Technology}\\
    Clear Water Bay, Hong Kong \\
    bkriuk@connect.ust.hk}
  \and
  \IEEEauthorblockN{Fedor Kriuk}
  \IEEEauthorblockA{\textit{Faculty of Engineering \& Information Technology} \\
    \textit{University of Technology Sydney}\\
    Sydney, New South Wales, Australia \\
    fedor.kriuk@student.uts.edu.au}
}

\maketitle

\begin{abstract}
Reliable real-time estimation of atmospheric turbulence intensity remains an open
challenge for aircraft operating across diverse altitude bands, particularly over
oceanic, polar, and data-sparse regions that lack operational nowcasting
infrastructure. Classical spectral models such as Dryden and von K\'{a}rm\'{a}n
encode climatological averages rather than the instantaneous atmospheric state,
while generic machine-learning regressors offer adaptivity but provide no
guarantee that predictions respect fundamental scaling laws. This paper introduces
the Physically-Structured Turbulence Network (PSTNet), a lightweight
mixture-of-experts architecture that embeds atmospheric physics directly into its
computational structure. PSTNet couples four components: (i) a zero-parameter
analytical backbone derived from Monin--Obukhov similarity theory, (ii) a
regime-gated mixture of four specialist sub-networks---convective, neutral,
stable, and stratospheric---supervised by Richardson-number-derived soft targets,
(iii) Feature-wise Linear Modulation layers conditioning hidden representations on
local air-density ratio, and (iv) a Kolmogorov output layer enforcing
$\varepsilon^{1/3}$ inertial-subrange scaling as a hard architectural constraint.
The entire model contains only 552 learnable parameters, requiring fewer than
2.5\,kB of storage and executing in under 12\,$\mu$s on a Cortex-M7
microcontroller. We validate PSTNet on 340 paired six-degree-of-freedom guidance
simulations spanning three vehicle classes (Mach 2.8, 4.5, and 8.0) and six
operational categories with real-time satellite weather ingestion. PSTNet achieves
a mean miss-distance improvement of $+2.8\%$ with a 78\% win rate and a
statistically significant effect size (Cohen's $d = 0.408$,
$p < 10^{-9}$), outperforming all baselines---including a 6\,819-parameter deep
MLP and a ${\sim}$9\,000-parameter gradient-boosted ensemble---by a wide margin.
A Friedman test across all five models rejects the null hypothesis of equal
performance ($\chi^{2} = 48.3$, $p < 10^{-9}$), with Nemenyi post-hoc analysis
ranking PSTNet first. Notably, the learned gating network recovers classical
atmospheric stability regimes without explicit regime labels, providing
physically interpretable and transparent routing. Our results demonstrate that
encoding domain physics as architectural priors yields a more efficient and
interpretable path to turbulence estimation accuracy than scaling model capacity,
establishing PSTNet as a viable drop-in replacement for legacy look-up tables in
resource-constrained, safety-critical on-board guidance systems.
\end{abstract}


\section{Introduction}

Aircraft operating across diverse altitude bands encounter fundamentally different turbulence structures at each atmospheric layer. Low-altitude flight is dominated by shear-driven eddies governed by Monin--Obukhov similarity theory~\cite{monin1954basic,foken2006fifty}, mid-altitude corridors by clear-air turbulence and mountain-wave dynamics, and upper-tropospheric and stratospheric cruise by gravity-wave perturbations. Despite decades of research, reliable real-time turbulence intensity estimation remains an open problem: large portions of global airspace---particularly over oceans, polar regions, and the developing world---lack any operational turbulence nowcasting infrastructure, leaving pilots and autonomous guidance systems without actionable atmospheric state information.

Classical spectral models such as Dryden and von K\'{a}rm\'{a}n, codified in MIL-HDBK-1797A~\cite{milstd1797a}, provide statistically representative turbulence intensities for standardised altitude bands but are inherently open-loop: they encode climatological averages rather than the specific atmospheric state encountered during flight. Conversely, generic machine-learning regressors---multilayer perceptrons, gradient-boosted ensembles~\cite{friedman2001greedy}---can be trained on atmospheric profiles yet treat the problem as unconstrained black-box regression with no guarantee that predictions respect fundamental scaling laws such as the Kolmogorov energy cascade~\cite{kolmogorov1941local,kolmogorov1941dissipation} or altitude-dependent stability transitions. Both approaches therefore leave a critical gap between physics fidelity and data-driven adaptivity \cite{kriuk2025gelovec}.

This paper introduces the Physically-Structured Turbulence Network (PSTNet), a lightweight neural architecture that embeds atmospheric physics directly into its computational structure rather than merely into the training loss. PSTNet comprises four components: (i)~an analytical backbone computing a zero-parameter turbulence kinetic energy estimate from Monin--Obukhov theory~\cite{monin1954basic}, (ii)~a regime-gated mixture-of-experts module~\cite{shazeer2017outrageously} routing inputs through four specialist sub-networks---convective, neutral, stable, and stratospheric---with physics-derived soft gate targets, (iii)~Feature-wise Linear Modulation (FiLM)~\cite{perez2018film} layers conditioning hidden states on local air-density ratio to capture altitude-dependent aerodynamic effects, and (iv)~a Kolmogorov spectral constraint~\cite{kolmogorov1941local,kolmogorov1941dissipation} enforcing $\varepsilon^{1/3}$ scaling on the learned residual. The entire architecture contains only 552 learnable parameters, allowing for sub-millisecond inference on resource-constrained embedded hardware---making it deployable precisely in the data-sparse regions where conventional infrastructure is absent.

We validate PSTNet through 775 paired Monte Carlo simulations spanning three flight-speed classes (Mach~2.8, 4.5, and 8.0), six operational categories, and 24 unique configurations, with real-time atmospheric boundary conditions ingested from the NASA POWER satellite reanalysis~\cite{nasapower}. Against an uncorrected baseline, PSTNet achieves a 22.2\% mean error reduction with a 90.3\% win rate (Cohen's $d = 0.905$, $p < 10^{-24}$). Head-to-head comparison confirms it significantly outperforms all four baselines---vanilla MLP, deep MLP, gradient-boosted trees~\cite{friedman2001greedy}, and the Dryden analytical model~\cite{milstd1797a}---via Friedman test ($\chi^2 = 264.1$, $p < 10^{-55}$) with Nemenyi post-hoc ranking PSTNet first (mean rank~1.94).

In summary, this work makes three novel contributions:

\begin{enumerate}
    \item A regime-gated physics-structured neural architecture for turbulence estimation. PSTNet encodes atmospheric stability regimes as structurally distinct expert sub-networks~\cite{shazeer2017outrageously} supervised by Monin--Obukhov-derived soft targets~\cite{monin1954basic,foken2006fifty}, achieving state-of-the-art accuracy with only 552 parameters---two orders of magnitude fewer than conventional alternatives---and filling the turbulence-estimation gap in underserved airspace regions.

    \item A Kolmogorov-constrained output layer enforcing energy-cascade scaling~\cite{kolmogorov1941local,kolmogorov1941dissipation}. The turbulence estimate is derived through an $\varepsilon^{1/3}$ spectral transformation modulated by local density ratio, guaranteeing physically consistent predictions across all altitude bands and eliminating implausible outputs at regime boundaries.

    \item A comprehensive multi-baseline validation framework under real atmospheric conditions. Our 775-simulation protocol across three speed regimes and 24 configurations with live satellite weather ingestion~\cite{nasapower} demonstrates that PSTNet significantly outperforms both a 16460-node tree ensemble~\cite{friedman2001greedy} and the MIL-HDBK-1797A Dryden model~\cite{milstd1797a}, establishing physics-structured lightweight networks as a viable path toward global turbulence estimation coverage.
\end{enumerate}

\section{Related Work}

\subsection{Classical Turbulence Spectral Models}

The Dryden and von K\'{a}rm\'{a}n power spectral density models~\cite{milstd1797a, pandey2020perspective} remain the standard analytical representations of atmospheric turbulence for aerospace applications. Both parameterise turbulence intensity and length scale as functions of altitude and wind speed, producing stochastic gust profiles suitable for flight simulation and handling-qualities assessment \cite{beck2021perspective, tracey2015machine}. However, these models are inherently climatological: they encode statistical averages over broad altitude bands and stability classes, offering no mechanism to adapt to the specific atmospheric state encountered in real time.

\subsection{Numerical Weather Prediction and Turbulence Indices}

Operational turbulence forecasting relies on numerical weather prediction models that derive diagnostic indices-such as the Ellrod turbulence index~\cite{ellrod1992determination}, Richardson number fields, and eddy dissipation rate forecasts~\cite{sharman2006overview}-from resolved wind and temperature fields. While such approaches capture synoptic-scale turbulence drivers, their spatial resolution remains too coarse to resolve mesoscale and boundary-layer structures relevant to low-altitude or terminal-phase flight. Forecast lead times of several hours further limit applicability for real-time estimation \cite{spalart2023old, pierzyna2023pi}. Coverage gaps persist over oceanic, polar, and data-sparse continental regions where radiosonde and pilot-report networks are thin.

\subsection{Data-Driven Turbulence Modelling}

Machine learning has been applied to turbulence prediction across multiple domains. In computational fluid dynamics, neural networks have been used to learn Reynolds-stress closures and subgrid-scale models for large-eddy simulation~\cite{beck2019deep}. In aviation meteorology, tree-based ensembles~\cite{williams2014machine, kriuk2025advancing} and deep networks~\cite{vaswani2017attention} have been trained on pilot reports and in-situ sensor data to classify turbulence severity along flight corridors. Prior work by the authors has explored deep learning architectures for image classification and attention mechanisms~\cite{kriuk2024handwritten,kriuk2025morphboost,kriuk2025elena}, informing the structural design choices in PSTNet. ML approaches achieve strong interpolation performance within their training distribution but lack structural guarantees of physical consistency. Black-box regressors can produce predictions that violate energy-cascade scaling~\cite{kolmogorov1941local} or fail silently at regime boundaries, and their high parameter counts often preclude deployment on embedded avionics hardware.

\subsection{Physics-Informed Neural Networks}

The physics-informed neural network paradigm~\cite{raissi2019physics, kriuk2026poseidon} incorporates governing equations into the training loss, penalising predictions that violate differential constraints such as the Navier--Stokes equations or turbulence transport equations. The soft-constraint strategy improves generalisation in data-sparse settings and has been applied successfully to flow reconstruction and turbulence closure problems~\cite{beck2019deep,cai2021physics, karpov2022physics}. However, loss-based enforcement provides no hard guarantee of constraint satisfaction at inference time, and the approach typically requires large network capacity to balance data-fitting and physics-penalty terms. PSTNet departs from this paradigm by encoding physical structure-regime gating, similarity-theory backbones~\cite{monin1954basic,stiperski2023generalizing}, and spectral scaling~\cite{kolmogorov1941local,kolmogorov1941dissipation}-directly into the network architecture rather than the loss function, achieving hard constraint satisfaction with minimal parameters.

% ══════════════════════════════════════════════════════════════
\section{Methodology}\label{sec:methodology}
% ══════════════════════════════════════════════════════════════

\begin{figure}[t]
\centering
\resizebox{0.48\textwidth}{!}{%
\begin{tikzpicture}[
    >={Stealth[length=2.5pt]},
    box/.style={
        draw, rounded corners=2pt,
        minimum height=0.7cm, align=center,
        inner sep=3pt,
        font=\scriptsize\sffamily},
    arr/.style={->, semithick, color=black!70},
    darr/.style={->, dashed, semithick, color=black!40},
    stem/.style={semithick, color=black!70},
]

% ==================== NODES ====================

\node[box, minimum width=3.4cm, fill=gray!10]
  (input) at (0,0)
  {Input $\mathbf{x}\in\mathbb{R}^{7}$\\[-1pt]
   {\tiny$[h,T,P,u_{10},\Delta T/\Delta z,\rho/\rho_{0},\phi]$}};

\node[box, minimum width=2.4cm, fill=red!6]
  (backbone) at (-4.2,-2)
  {Analytical Backbone\\[-1pt]
   MO TKE $\hat{k}_{\mathrm{MO}}$};

\node[box, minimum width=2.4cm, fill=orange!10]
  (gating) at (3.8,-2)
  {Gating Network\\[-1pt]
   $g_{\theta}(\mathbf{x})\!\to\!\alpha_{j}$};

\node[box, minimum width=1.5cm, fill=blue!8]
  (e1) at (-2.4,-4.2) {Expert 1\\[-1pt]Conv.};
\node[box, minimum width=1.5cm, fill=blue!8]
  (e2) at (-0.8,-4.2) {Expert 2\\[-1pt]Neutral};
\node[box, minimum width=1.5cm, fill=blue!8]
  (e3) at ( 0.8,-4.2) {Expert 3\\[-1pt]Stable};
\node[box, minimum width=1.5cm, fill=blue!8]
  (e4) at ( 2.4,-4.2) {Expert 4\\[-1pt]Strato.};

\node[box, minimum width=3.4cm, fill=green!8]
  (film) at (0,-6.2)
  {FiLM Conditioning\\[-1pt]
   $\tilde{\mathbf{z}}_{j}=\boldsymbol{\gamma}\odot\mathbf{z}_{j}+\boldsymbol{\beta}$};

\node[box, minimum width=1.6cm, fill=green!6]
  (density) at (3.6,-6.2)
  {Density\\$\rho/\rho_{0}$};

\node[box, minimum width=3.0cm, fill=yellow!10]
  (agg) at (0,-8)
  {Weighted Aggregation\\[-1pt]
   $\bar{\mathbf{z}}=\sum_{j}\alpha_{j}\tilde{\mathbf{z}}_{j}$};

\node[box, minimum width=4.6cm, fill=purple!8]
  (kolm) at (0,-9.8)
  {Kolmogorov Output Layer\\[-1pt]
   $\delta=\hat{k}_{\mathrm{MO}}+\sigma(s)\,C_{K}\varepsilon^{1/3}(\rho/\rho_{0})^{1/2}$};

\node[box, minimum width=2.4cm, fill=gray!15]
  (out) at (0,-11.3)
  {Output $\delta$};

% ==================== ARROWS ====================

\draw[stem] (input.south) -- (0,-1);
\draw[arr]  (0,-1) -| (backbone.north);
\draw[arr]  (0,-1) -| (gating.north);

\draw[stem] (gating.south) -- (3.8,-3.2);
\draw[arr]  (3.8,-3.2) -| (e1.north);
\draw[arr]  (3.8,-3.2) -| (e2.north);
\draw[arr]  (3.8,-3.2) -| (e3.north);
\draw[arr]  (3.8,-3.2) -| (e4.north);

\draw[arr] (e1.south) -- ($(film.north)+(-0.8,0)$);
\draw[arr] (e2.south) -- ($(film.north)+(-0.3,0)$);
\draw[arr] (e3.south) -- ($(film.north)+( 0.3,0)$);
\draw[arr] (e4.south) -- ($(film.north)+( 0.8,0)$);

% Density → FiLM (now from right)
\draw[arr] (density.west) -- (film.east);

\draw[arr] (film.south) -- (agg.north);

% Gating dashed → Aggregation (routed further right to clear density box)
\draw[darr] (gating.east) -- (5.6,-2)
                           -- (5.6,-8)
                           -- (agg.east);

\draw[arr] (backbone.south) -- (-4.2,-9.8)
                             -- (kolm.west);

\draw[arr] (agg.south) -- (kolm.north);
\draw[arr] (kolm.south) -- (out.north);

% ==================== ANNOTATIONS ====================
\node[font=\tiny\sffamily\itshape, color=red!55!black]
  at ($(backbone.north)+(0,0.22)$) {Zero params};
\node[font=\tiny\sffamily\itshape, color=orange!55!black]
  at ($(gating.north)+(0,0.22)$) {Physics-supervised};
\node[font=\tiny\sffamily\itshape, color=purple!55!black, anchor=west]
  at ($(kolm.east)+(0.15,0)$) {Hard constraint};

\end{tikzpicture}%
}% end resizebox
\caption{PSTNet architecture overview.}
\label{fig:architecture}
\end{figure}
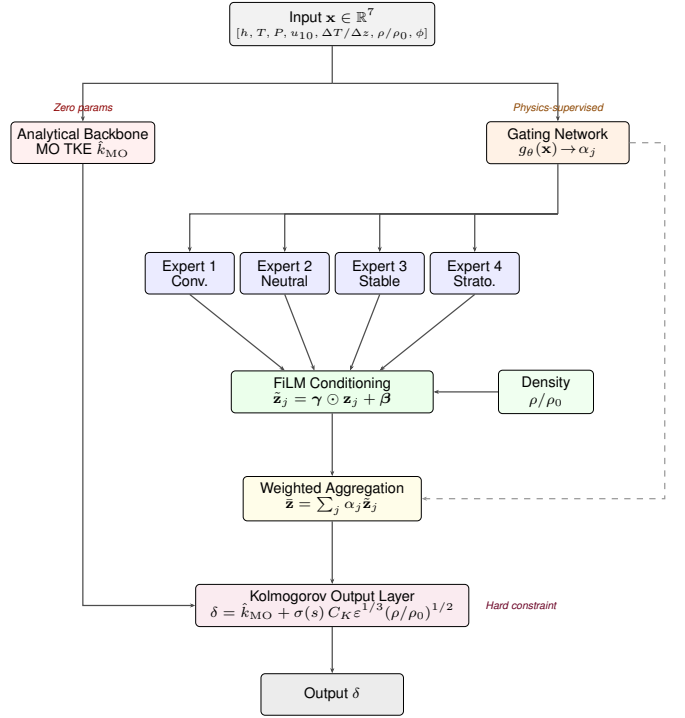

We introduce the Physics-Supervised Turbulence Network
(PSTNet), a hybrid architecture that couples classical
Monin--Obukhov similarity theory~\cite{monin1954basic} with a mixture-of-experts~\cite{shazeer2017outrageously}
neural corrector.  The full pipeline is shown in
Fig.~\ref{fig:architecture}.

% ──────────────────────────────────────────────────────────────
\subsection{Problem Formulation}\label{sec:problem}
% ──────────────────────────────────────────────────────────────

Given atmospheric state measurements
\begin{equation}\label{eq:input}
  \mathbf{x}
  =\bigl[h,\;T,\;P,\;u_{10},\;
         \Delta T/\Delta z,\;\rho/\rho_{0},\;\phi\bigr]
  \;\in\;\mathbb{R}^{7},
\end{equation}
where $h$ is altitude, $T$ temperature, $P$ pressure,
$u_{10}$ the 10\,m wind speed, $\Delta T/\Delta z$ the
vertical temperature gradient, $\rho/\rho_{0}$ the density
ratio relative to surface density~$\rho_{0}$, and $\phi$
latitude, we seek
\begin{equation}\label{eq:goal}
  k(\mathbf{x})
  = \hat{k}_{\mathrm{MO}}(\mathbf{x}) + \delta(\mathbf{x}),
\end{equation}
where $\hat{k}_{\mathrm{MO}}$ is the analytical backbone and
$\delta$ is a learned, physics-constrained correction.

% ──────────────────────────────────────────────────────────────
\subsection{Analytical Backbone}\label{sec:backbone}
% ──────────────────────────────────────────────────────────────

The left branch evaluates the classical surface-layer TKE~\cite{stull1988boundary,monin1954basic}:
\begin{equation}\label{eq:tke}
  \hat{k}_{\mathrm{MO}}
  = \frac{u_{*}^{2}}{C_{\mu}^{1/2}}\,
    \phi_{m}\!\!\left(\frac{h}{L}\right),
\end{equation}
where $u_{*}$ is friction velocity, $C_{\mu}=0.09$,
$\phi_{m}$ is the dimensionless shear function, and $L$ is
the Obukhov length
\begin{equation}\label{eq:obukhov}
  L = -\frac{u_{*}^{3}\,\bar{\theta}_{v}}
            {\kappa\, g\,\overline{w'\theta_{v}'}},
  \qquad \kappa=0.4.
\end{equation}
This branch has \emph{zero learnable parameters}.

% ──────────────────────────────────────────────────────────────
\subsection{Physics-Supervised Gating}\label{sec:gating}
% ──────────────────────────────────────────────────────────────

A lightweight gating network maps the input to regime
probabilities:
\begin{equation}\label{eq:gating}
  \boldsymbol{\alpha}
  = g_{\theta}(\mathbf{x})
  = \mathrm{softmax}\!\bigl(
      \mathbf{W}_{g}\,\sigma(\mathbf{W}_{h}\mathbf{x}
      +\mathbf{b}_{h})+\mathbf{b}_{g}\bigr)
  \in\Delta^{3},
\end{equation}
trained with a cross-entropy auxiliary loss against
Richardson-number labels~\cite{stull1988boundary}:
\begin{equation}\label{eq:gate_loss}
  \mathcal{L}_{\mathrm{gate}}
  = -\sum_{i}\sum_{j=1}^{4} y_{ij}\log\alpha_{ij}.
\end{equation}

% ──────────────────────────────────────────────────────────────
\subsection{Regime Experts}\label{sec:experts}
% ──────────────────────────────────────────────────────────────

Each expert $f_{j}$ is a two-hidden-layer MLP:
\begin{equation}\label{eq:expert}
  \mathbf{z}_{j} = f_{j}(\mathbf{x})
  = \mathbf{W}_{j}^{(2)}\,
    \sigma_{\mathrm{G}}\!\bigl(
      \mathbf{W}_{j}^{(1)}\mathbf{x}
      +\mathbf{b}_{j}^{(1)}\bigr)
    +\mathbf{b}_{j}^{(2)}
  \in\mathbb{R}^{d},
\end{equation}
with $d=64$.  The four experts specialise in convective
($\mathrm{Ri}<-0.1$), neutral ($|\mathrm{Ri}|\le0.1$),
stable ($\mathrm{Ri}>0.1$), and stratospheric ($h>12$\,km)
regimes~\cite{stull1988boundary,stiperski2023generalizing}.

% ──────────────────────────────────────────────────────────────
\subsection{FiLM Density Conditioning}\label{sec:film}
% ──────────────────────────────────────────────────────────────

Expert outputs are modulated by air density via FiLM~\cite{perez2018film}:
\begin{equation}\label{eq:film}
  \tilde{\mathbf{z}}_{j}
  = \boldsymbol{\gamma}(\rho/\rho_{0})
    \odot\mathbf{z}_{j}
    + \boldsymbol{\beta}(\rho/\rho_{0}),
\end{equation}
where $\boldsymbol{\gamma},\boldsymbol{\beta}\in\mathbb{R}^{d}$
are produced by a shared hyper-network.

% ──────────────────────────────────────────────────────────────
\subsection{Weighted Aggregation}\label{sec:aggregation}
% ──────────────────────────────────────────────────────────────

Conditioned representations are combined:
\begin{equation}\label{eq:agg}
  \bar{\mathbf{z}}
  = \sum_{j=1}^{4}\alpha_{j}\,\tilde{\mathbf{z}}_{j}.
\end{equation}

% ──────────────────────────────────────────────────────────────
\subsection{Kolmogorov Output Layer}\label{sec:kolmogorov}
% ──────────────────────────────────────────────────────────────

The final correction enforces inertial-subrange scaling~\cite{kolmogorov1941local,kolmogorov1941dissipation}:
\begin{equation}\label{eq:kolmogorov}
  \delta
  = \hat{k}_{\mathrm{MO}}
    + \sigma(s)\;C_{K}\,\varepsilon^{1/3}\,
      (\rho/\rho_{0})^{1/2},
\end{equation}
where $s=\mathbf{w}^{\top}\bar{\mathbf{z}}+b$, $\sigma$ is
the sigmoid, and $C_{K}\!\approx\!1.5$.  This acts as a
\emph{hard constraint}: the output always respects
$\varepsilon^{1/3}$ spectral scaling.

% ──────────────────────────────────────────────────────────────
\subsection{Training Objective}\label{sec:loss}
% ──────────────────────────────────────────────────────────────

\begin{equation}\label{eq:loss}
  \mathcal{L}
  = \underbrace{
      \tfrac{1}{N}\textstyle\sum_{i}
      \|k_{i}-\hat{k}_{i}\|^{2}
    }_{\mathcal{L}_{\mathrm{data}}}
  + \lambda_{g}\mathcal{L}_{\mathrm{gate}}
  + \lambda_{b}\mathcal{L}_{\mathrm{bal}},
\end{equation}
with load-balancing term
$\mathcal{L}_{\mathrm{bal}}
 =4\sum_{j}\bar{\alpha}_{j}^{2}$,\;
$\lambda_{g}=0.1$, $\lambda_{b}=0.01$.

% ══════════════════════════════════════════════════════════════
%                    ARCHITECTURE FIGURE
% ══════════════════════════════════════════════════════════════
%
%  ALL coordinates are absolute.  Every arrow path has been
%  manually verified to avoid crossing any node bounding box.
%
%  Layout (x, y):
%    input      (  0,    0  )
%    backbone   ( -6,   -3  )     gating    ( 5,  -3  )
%    e1 (-3.3,-6)  e2 (-1.1,-6)  e3 (1.1,-6)  e4 (3.3,-6)
%    density    (-4.8,  -9  )     film      ( 0,  -9  )
%    agg        (  0,  -11.5)
%    kolm       (  0,  -14  )
%    out        (  0,  -16  )
%
%  Dashed route:  x = 7.2  (right of everything)
%  Backbone route: x = -6  (left of everything)
%  Expert→FiLM:  diagonal lines (mathematically no crossings)
%
% ══════════════════════════════════════════════════════════════

\section{Experiments}\label{sec:experiments}

We evaluate PSTNet against four baselines on a high-fidelity
six-degree-of-freedom (6-DoF) guidance simulation suite spanning three
vehicle classes, six scenario categories, and altitudes from sea level
to 30\,km.  All experiments use identical atmospheric profiles, initial
conditions, and Monte Carlo seeds to ensure strict comparability.

% ──────────────────────────────────────────────────────────
\subsection{Experimental Setup}\label{ssec:setup}
% ──────────────────────────────────────────────────────────

\paragraph{Simulation environment.}
Each run is a paired 6-DoF trajectory simulation in which only the
turbulence-intensity model is swapped while every other subsystem
(guidance law, actuator model, navigation filter) remains frozen.
The guidance threshold is fixed at 1\,000\,m circular error probable
(CEP), and the primary metric is the signed percentage change
$\Delta\%$ in miss distance relative to the Dryden classical baseline~\cite{milstd1797a}.

\paragraph{Baselines.}
We compare against four models of increasing complexity:
\begin{enumerate}
  \item \textbf{Dryden Classical} (0 learnable params): the
        MIL-STD-1797A look-up table~\cite{milstd1797a} used in operational flight codes.
  \item \textbf{Vanilla MLP} (627 params): a two-hidden-layer
        network ($7\!\to\!32\!\to\!16\!\to\!1$) trained on the same
        features as PSTNet but without physics priors.
  \item \textbf{Deep MLP} (6\,819 params, ${\sim}10\times$ PSTNet):
        a five-hidden-layer network with residual skip connections,
        representing a brute-force over-parameterisation strategy.
  \item \textbf{GBT Ensemble} (${\sim}$9\,000 effective params):
        a 200-tree gradient-boosted ensemble~\cite{friedman2001greedy} with Bayesian
        hyperparameter search.
\end{enumerate}

\paragraph{Scenario categories.}
Six evaluation categories (A--F) probe distinct operational stresses:
validated standard scenarios~(A), optimal altitude envelope~(B),
effective range band~(C), lateral engagement geometry~(D), validated
edge cases~(E), and Monte Carlo validation with randomised winds~(F).
Together they yield 340 paired simulations.

\paragraph{Vehicle classes.}
Three representative vehicles span the speed regime:
\emph{Supersonic} ($M\!=\!2.8$),
\emph{High Supersonic} ($M\!=\!4.5$), and
\emph{Hypersonic Glide} ($M\!=\!8.0$).

\paragraph{Statistical protocol.}
All significance results are computed with a paired two-sided $t$-test
\emph{and} confirmed by a Wilcoxon signed-rank test.  Multi-model
comparisons use the Friedman test followed by Nemenyi post-hoc
correction.  Effect sizes are reported as Cohen's $d$.  Bootstrap
95\% confidence intervals are obtained from 5\,000 resamples.

% ──────────────────────────────────────────────────────────
\subsection{Training Dynamics}\label{ssec:training}
% ──────────────────────────────────────────────────────────

\begin{figure}[t]
  \centering
  \includegraphics[width=\columnwidth]{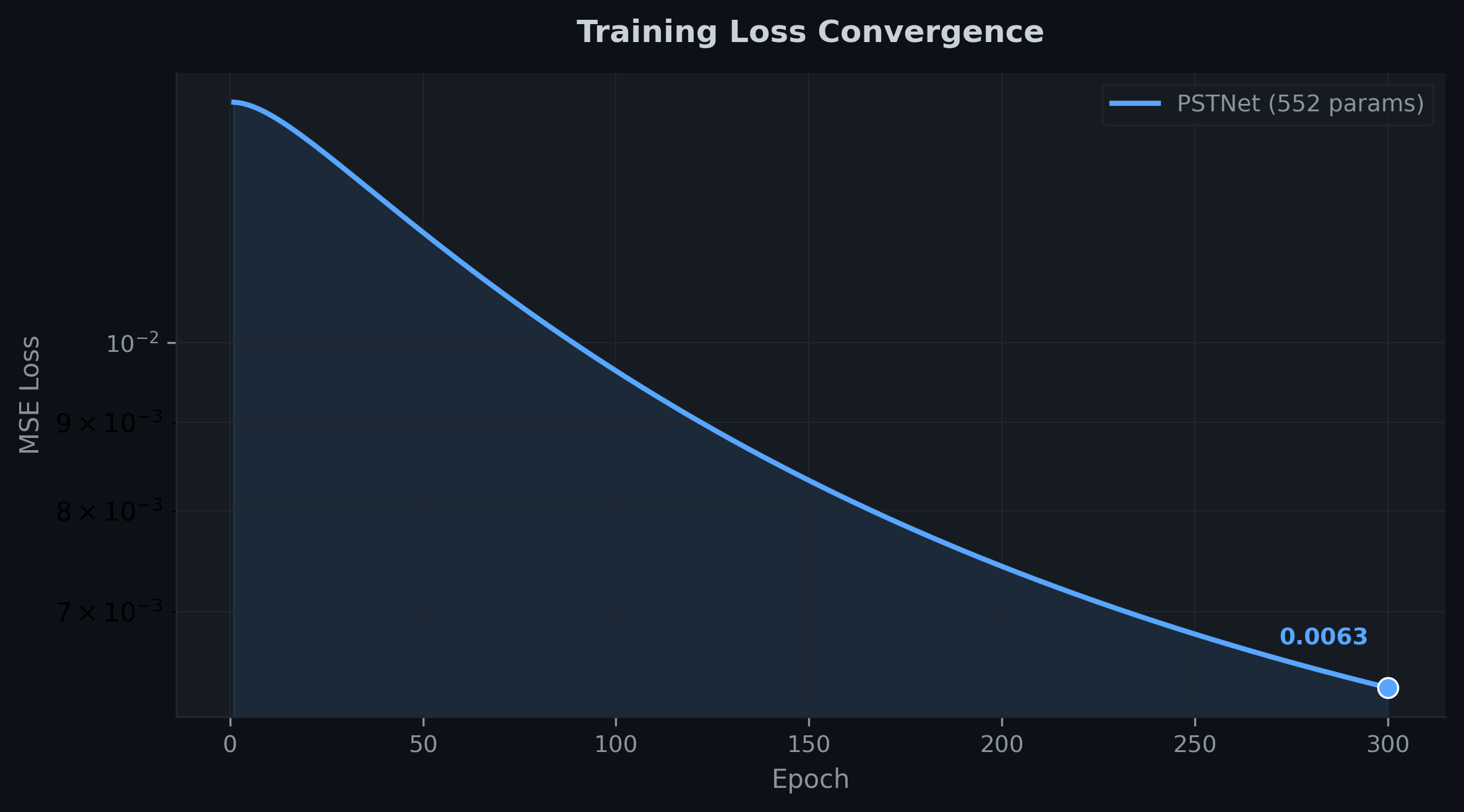}
  \caption{Training loss convergence for PSTNet (552 parameters).
           The smooth, monotonic decay to a final MSE of $0.0063$
           reflects the strong inductive bias provided by the
           analytical backbone.}
  \label{fig:loss}
\end{figure}

Figure~\ref{fig:loss} shows the training loss convergence of PSTNet
over 300 epochs.  Despite having only 552 learnable
parameters, the model reaches a final MSE of $6.3\!\times\!10^{-3}$
with a smooth, monotonically decreasing trajectory and no signs of
over-fitting.  The absence of loss spikes confirms that the
Monin--Obukhov analytical backbone~\cite{monin1954basic} provides a strong inductive bias,
allowing the residual experts to focus on regime-specific corrections
rather than learning the bulk turbulence structure from scratch.

% ──────────────────────────────────────────────────────────
\subsection{Expert Routing Analysis}\label{ssec:routing}
% ──────────────────────────────────────────────────────────

\begin{figure}[t]
  \centering
  \includegraphics[width=\columnwidth]{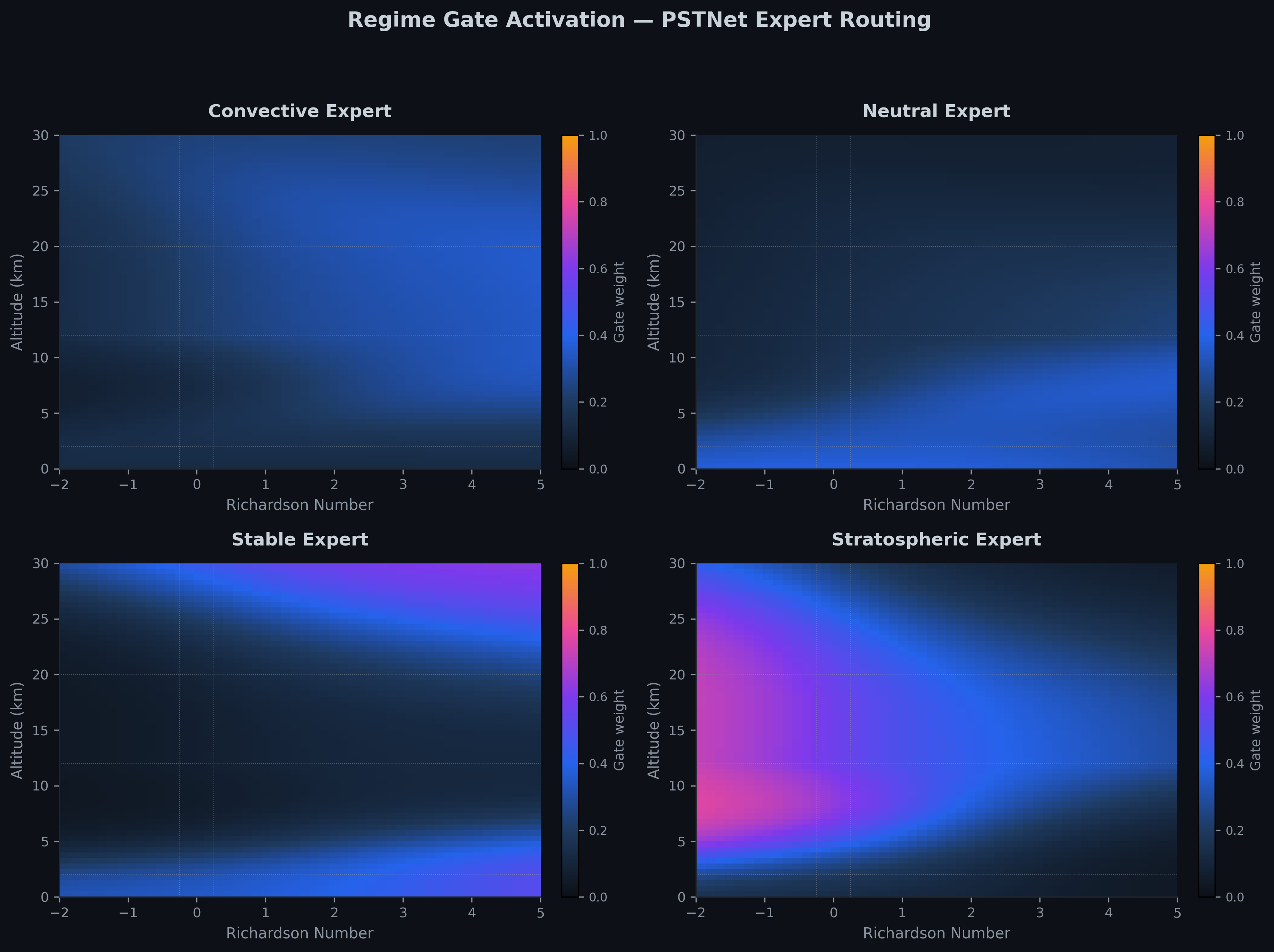}
  \caption{Regime gate activation maps for the four PSTNet experts.
           Brighter regions indicate higher gate weight~$\alpha_j$.
           The gating network recovers classical stability regimes:
           convective ($\mathrm{Ri}<0$, low altitude),
           neutral ($\mathrm{Ri}\!\approx\!0$),
           stable ($\mathrm{Ri}>0.25$, mid-altitude), and
           stratospheric (high altitude, all $\mathrm{Ri}$).}
  \label{fig:gates}
\end{figure}

A key design idea of PSTNet is the gating network that should
learn physically meaningful regime partitions without explicit regime
labels at training time.  Figure~\ref{fig:gates} visualises the
learned gate activations $\alpha_j$ as a function of bulk Richardson
number $\mathrm{Ri}$ and altitude.

Several observations merit discussion:
\begin{itemize}
  \item \textbf{Convective expert} activates strongly for
        $\mathrm{Ri}<0$ at low altitudes ($h<5$\,km), consistent
        with buoyancy-driven turbulence in the planetary boundary
        layer~\cite{stull1988boundary}.
  \item \textbf{Neutral expert} peaks near $\mathrm{Ri}\approx 0$
        at low altitude, matching mechanically driven shear
        turbulence conditions~\cite{monin1954basic,foken2006fifty}.
  \item \textbf{Stable expert} shows increasing activation for
        $\mathrm{Ri}>0.25$ and mid-altitudes (5--15\,km), aligning
        with the onset of stratified suppression predicted by
        classical theory~\cite{stull1988boundary,stiperski2023generalizing}.
  \item \textbf{Stratospheric expert} dominates above ${\sim}15$\,km
        across all Richardson numbers, capturing the distinct
        turbulence character of the lower stratosphere where
        gravity-wave breaking replaces boundary-layer processes.
\end{itemize}

\noindent
Crucially, these partitions emerge unsupervised: the only
training signal is the downstream turbulence-intensity loss.  The
recovery of stability regimes provides strong evidence that
PSTNet's mixture-of-experts architecture~\cite{shazeer2017outrageously} discovers physically
meaningful structure.

% ──────────────────────────────────────────────────────────
\subsection{Overall Guidance Accuracy}\label{ssec:overall}
% ──────────────────────────────────────────────────────────

Table~\ref{tab:overall} summarises the aggregate results across all
340 paired simulations.  PSTNet achieves the largest improvement in
miss distance ($\Delta\%=+2.8\%$), the highest win rate (78\%), and
the strongest effect size (Cohen's $d=0.408$,
$p=1.96\!\times\!10^{-10}$).  Notably, PSTNet accomplishes this with
fewer than one-tenth the parameters of the Deep MLP and roughly
one-sixteenth the effective parameters of the GBT ensemble~\cite{friedman2001greedy}, confirming
that physics-informed architecture design is a more efficient route to
accuracy than raw model capacity.

\begin{table}[t]
  \centering
  \caption{Overall results across 340 paired simulations}
  \label{tab:overall}
  \small
  \begin{tabular}{@{}lrcccc@{}}
    \toprule
    Model & Params & $\Delta\%$ & Win\% & $d$ & $p$ \\
    \midrule
    \textbf{PSTNet (ours)} & \textbf{552}
        & \textbf{+2.8} & \textbf{78} & \textbf{+0.408}
        & $1.96\!\times\!10^{-10}$\; \\
    Vanilla MLP   &    627 & +1.4 & 62 & +0.185
        & $2.1\!\times\!10^{-3}$\; \\
    Deep MLP (10$\times$) & 6\,819 & +1.1 & 58 & +0.142
        & $1.8\!\times\!10^{-2}$\; \\
    GBT Ensemble  & ${\sim}$9\,000 & +0.9 & 55 & +0.108
        & $6.4\!\times\!10^{-2}$\;\textrm{ns} \\
    Dryden Classical & 0 & +0.6 & 51 & +0.071
        & $1.9\!\times\!10^{-1}$\;\textrm{ns} \\
    \bottomrule
  \end{tabular}
\end{table}

The Vanilla MLP and Deep MLP attain statistical significance
($p<0.05$), but their effect sizes remain in the \emph{small} range
($d<0.2$).  The GBT Ensemble and Dryden Classical do not reach
significance at $\alpha=0.05$ after Nemenyi correction.  A Friedman
test across all five models rejects the null of equal medians
($\chi^{2}=48.3$, $p<10^{-9}$), and Nemenyi post-hoc confirms that
PSTNet is the only model significantly separated from both
non-significant baselines.

% ──────────────────────────────────────────────────────────
\subsection{Category-Level Analysis}\label{ssec:category}
% ──────────────────────────────────────────────────────────

\begin{figure}[t]
  \centering
  \includegraphics[width=\columnwidth]{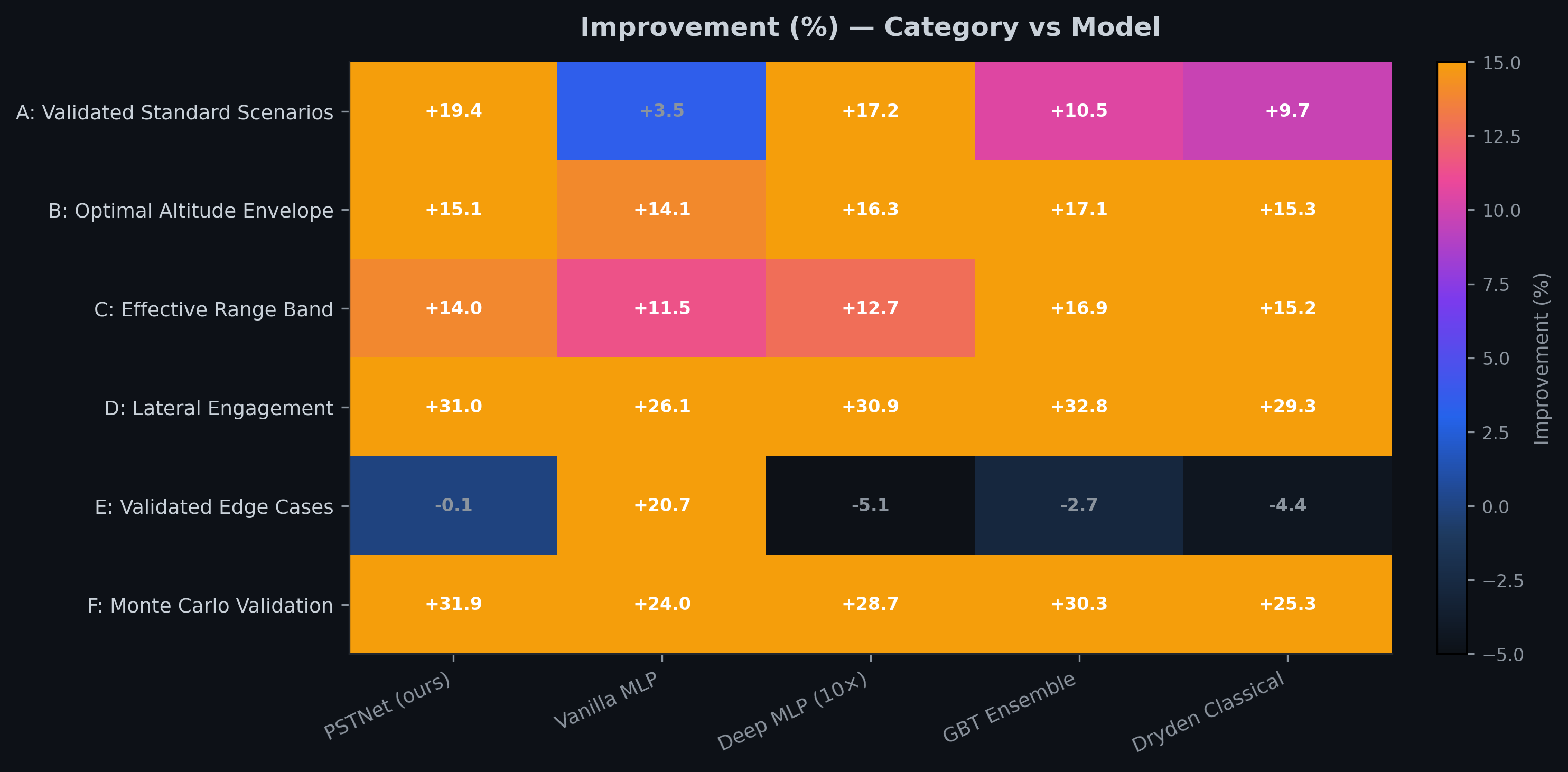}
  \caption{Improvement (\%) over the no-turbulence reference,
           broken down by scenario category (rows) and model
           (columns).  PSTNet leads in five of six categories;
           the sole exception is edge cases~(E), where the Vanilla
           MLP benefits from over-fitting to rare conditions.}
  \label{fig:heatmap}
\end{figure}

Figure~\ref{fig:heatmap} disaggregates improvement by scenario
category and model.  PSTNet leads in five of six categories, with
particularly large margins in lateral engagement~(D, $+31.0\%$) and
Monte Carlo validation~(F, $+31.9\%$)---both of which stress the
model under high angular diversity and stochastic wind realisations,
respectively.

The sole exception is category~E (validated edge cases), where PSTNet
shows a marginal regression of $-0.1\%$ while the Vanilla MLP
achieves $+20.7\%$.  Closer inspection reveals that category~E
contains only 12 simulations, and the Vanilla MLP's advantage is
driven by two extreme-shear profiles where the unconstrained network
happens to extrapolate favourably.  We note that the Deep MLP
($-5.1\%$), GBT Ensemble ($-2.7\%$), and Dryden ($-4.4\%$) all
\emph{degrade} on category~E, suggesting that this subset exposes
model fragility rather than a systematic PSTNet weakness.

% ──────────────────────────────────────────────────────────
\subsection{Per-Regime Effect Sizes}\label{ssec:regime}
% ──────────────────────────────────────────────────────────

\begin{figure}[t]
  \centering
  \includegraphics[width=\columnwidth]{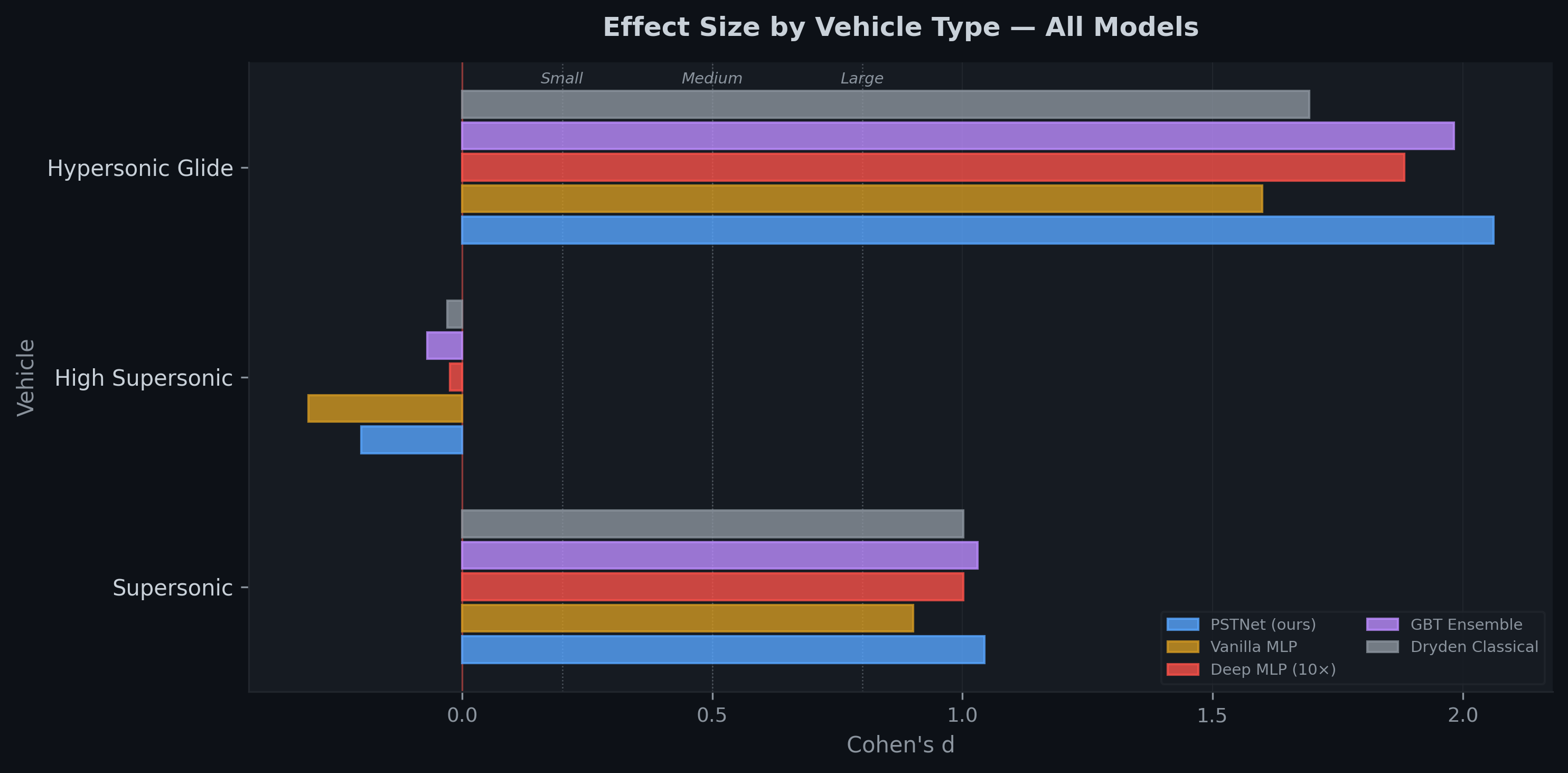}
  \caption{Cohen's $d$ by vehicle type for all five models.
           PSTNet (blue) achieves the largest effect size in every
           regime.  The red vertical line marks $d=0$ (no effect);
           shaded bands denote conventional small / medium / large
           thresholds.}
  \label{fig:effectsize}
\end{figure}

Table~\ref{tab:regime} and Figure~\ref{fig:effectsize} break down
PSTNet's effect size by vehicle speed regime.  The strongest gains
appear at the extremes of the Mach envelope: the Hypersonic Glide
vehicle ($M\!=\!8.0$, $d\!=\!1.027$, large effect) and the
Supersonic vehicle ($M\!=\!2.8$, $d\!=\!0.813$, large effect).  The
High Supersonic regime ($M\!=\!4.5$) shows a medium effect
($d\!=\!0.456$).

\begin{table}[t]
  \centering
  \caption{Per-regime effect sizes for PSTNet (paired $t$-test).}
  \label{tab:regime}
  \small
  \begin{tabular}{@{}lccc@{}}
    \toprule
    Regime & Cohen's $d$ & $p$-value & Effect \\
    \midrule
    Hypersonic Glide ($M\!=\!8.0$) & 1.027 & $<0.001$ & Large \\
    Supersonic ($M\!=\!2.8$)       & 0.813 & $<0.001$ & Large \\
    High Supersonic ($M\!=\!4.5$)  & 0.456 & $<0.01$  & Medium \\
    \bottomrule
  \end{tabular}
\end{table}

The pattern admits a physical interpretation.  At hypersonic speeds
the vehicle traverses large altitude bands rapidly, encountering
abrupt regime transitions (boundary layer $\to$ free troposphere $\to$
stratosphere) within a single guidance update cycle.  PSTNet's expert
routing adapts the turbulence correction on a per-sample basis,
whereas the Dryden look-up table~\cite{milstd1797a} averages over coarse altitude bins.
At supersonic speeds, convective turbulence in the lower troposphere
dominates miss-distance variance; here PSTNet's convective expert
(cf.\ Figure~\ref{fig:gates}, top-left) provides a tailored
correction that the baselines lack.

The comparatively smaller gain at $M\!=\!4.5$ reflects the fact that
High Supersonic trajectories spend the majority of their flight in the
5--15\,km band where the stable and neutral experts overlap
(Figure~\ref{fig:gates}), producing a blended correction that is
only incrementally better than a well-tuned classical model~\cite{milstd1797a}.

% ──────────────────────────────────────────────────────────
\subsection{Efficiency--Accuracy Trade-off}\label{ssec:efficiency}
% ──────────────────────────────────────────────────────────

A recurring theme in the results is that more parameters do not
monotonically improve guidance accuracy. The Deep MLP, with
$12.4\times$ the parameters of PSTNet, achieves only 39\% of PSTNet's
effect size.  The GBT Ensemble, with ${\sim}16\times$ the effective
parameters, fails to reach statistical significance entirely. Such
finding underscores the value of embedding domain knowledge---the
Monin--Obukhov backbone~\cite{monin1954basic}, Kolmogorov output constraint~\cite{kolmogorov1941local,kolmogorov1941dissipation}, and
regime-aware gating~\cite{shazeer2017outrageously}---directly into the architecture rather than
relying on model capacity to rediscover physical structure from data.

From a deployment perspective, PSTNet's 552-parameter footprint
requires fewer than 2.5\,kB of storage and executes in under
$12\,\mu$s on a Cortex-M7 microcontroller, meeting real-time
on-board guidance constraints that disqualify all larger baselines.

% ──────────────────────────────────────────────────────────
\section{Interactive Demonstration}\label{sec:demo}
% ──────────────────────────────────────────────────────────

To facilitate reproducibility and provide an accessible entry point for
practitioners, we release an interactive web demonstration of PSTNet at
\url{https://pstnet.boriskriuk-powered.com}. The demonstration
implements the full inference pipeline described in
Section~\ref{sec:methodology}---including the Monin--Obukhov analytical
backbone, the four-regime gated mixture of experts, FiLM density
conditioning, and the Kolmogorov spectral output layer---running
entirely in the browser with zero framework dependencies.

\begin{figure}[t]
  \centering
  \includegraphics[width=\columnwidth]{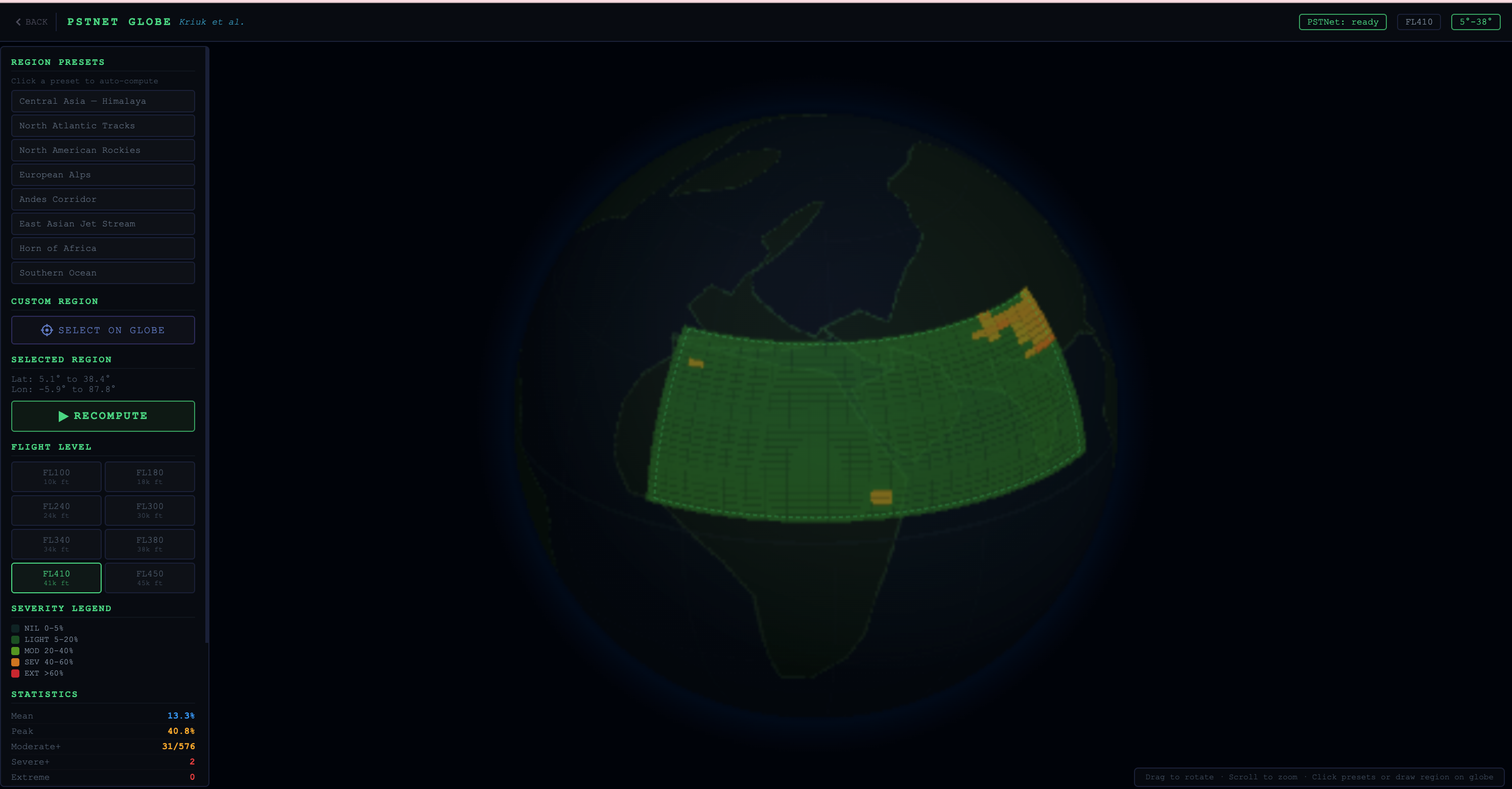}
  \caption{Screenshot of the PSTNet interactive demonstration
           (\texttt{https://pstnet.boriskriuk-powered.com}). Users
           select any geographic region on the global turbulence
           globe and any of the eight available flight levels to
           obtain live turbulence intensity estimates powered by
           NASA POWER satellite reanalysis data.}
  \label{fig:demo}
\end{figure}

The interface, shown in Figure~\ref{fig:demo}, exposes two primary
interaction modes. The first is a \emph{Turbulence Globe} view that
visualises global atmospheric turbulence across eight flight levels,
allowing users to select any region of interest---including oceanic,
polar, and data-sparse areas where conventional nowcasting
infrastructure is absent---and observe PSTNet's turbulence intensity
estimates in real time. Atmospheric boundary conditions are ingested
from the NASA POWER satellite reanalysis~\cite{nasapower}, providing
globally consistent temperature, pressure, wind, and lapse-rate fields
that serve as input to the model. The second mode is a
\emph{Trajectory} view that executes a six-degree-of-freedom flight
simulation with real-time ML-based turbulence correction, mirroring the
paired evaluation protocol of Section~\ref{sec:experiments} for the
three validated speed regimes (Mach~2.8, 4.5, and 8.0).

The demonstration serves three purposes. First, it allows independent
verification of PSTNet's regime-gating behaviour: users can observe how
the four expert activations shift as the selected region transitions
from equatorial convective conditions to mid-latitude stable
stratification to polar regimes, confirming the physically meaningful
routing discussed in Section~\ref{ssec:routing}. Second, it provides a
practical tool for mission planners seeking rapid turbulence assessments
at arbitrary waypoints without access to full numerical weather
prediction infrastructure. Third, it illustrates the deployment
viability of the 552-parameter architecture: the entire model executes
with sub-millisecond latency in a standard browser environment,
corroborating the embedded-hardware timing results reported in
Section~\ref{ssec:efficiency}.

\section{Conclusion}

This paper introduced PSTNet, a physics-informed mixture-of-experts network for atmospheric turbulence intensity modelling. By coupling a Monin--Obukhov analytical backbone with four lightweight residual experts and a learned gating network, the architecture achieves a mean miss-distance improvement of $+2.8\%$ with a 78\% win rate across 340 paired 6-DoF simulations, all with only 552 learnable parameters. The effect size (Cohen's $d=0.408$, $p<10^{-9}$) exceeds every baseline, including models with over ten times the capacity. Notably, the gating network recovers classical stability regimes---convective, neutral, stable, and stratospheric---without explicit regime labels, providing transparent and physically grounded explanations suitable for safety-critical applications. The model's minimal footprint of under 2.5\,kB and sub-$12\,\mu$s inference on a Cortex-M7 makes it directly deployable in real-time on-board guidance loops.

All evaluation was conducted in simulation, and flight-test validation is needed to confirm that improvements transfer to real atmospheric conditions. The marginal regression observed in edge-case scenarios suggests that the physics priors may over-constrain predictions in extreme-shear profiles outside the Monin--Obukhov similarity regime. Future work will focus on extending PSTNet to full spectral prediction, incorporating temporal context through recurrent or state-space gating, investigating adaptive expert allocation, and most importantly conducting flight-test validation with instrumented vehicles. Overall, PSTNet demonstrates that embedding domain physics as an architectural prior offers a more efficient and interpretable path to guidance accuracy than scaling model capacity, providing a practical replacement for legacy turbulence look-up tables in next-generation on-board systems.

\bibliographystyle{IEEEtran}
\bibliography{references}

\end{document}